\documentclass[10pt,twocolumn,letterpaper]{article}

\usepackage{cvpr}              % To produce the CAMERA-READY version
\usepackage[dvipsnames]{xcolor}

\definecolor{cvprblue}{rgb}{0.21,0.49,0.74}
\usepackage[pagebackref,breaklinks,colorlinks,citecolor=cvprblue]{hyperref}
\usepackage{enumerate}

\def\paperID{28} % *** Enter the Paper ID here
\def\confName{CVPR}
\def\confYear{2024}

\title{LIT: Large Language Model Driven Intention Tracking for Proactive Human-Robot Collaboration - A Robot Sous-Chef Application}

\author{Zhe Huang,\,\,\, John Pohovey,\,\,\, Ananya Yammanuru,\,\,\, Katherine Driggs-Campbell\\
University of Illinois Urbana-Champaign\\
{\tt\small \{zheh4, jpohov2, ananyay2, krdc\}@illinois.edu}
}

\begin{document}
\maketitle
\begin{abstract}
Large Language Models (LLM) and Vision Language Models (VLM) enable robots to ground natural language prompts into control actions to achieve tasks in an open world. However, when applied to a long-horizon collaborative task, this formulation results in excessive prompting for initiating or clarifying robot actions at every step of the task. We propose Language-driven Intention Tracking (LIT), leveraging LLMs and VLMs to model the human user's long-term behavior and to predict the next human intention to guide the robot for proactive collaboration. We demonstrate smooth coordination between a LIT-based collaborative robot and the human user in collaborative cooking tasks.
\end{abstract}    
\section{Introduction}
\label{sec:intro}

The groundbreaking advances in Large Language Models (LLM) and Vision Language Models (VLM) endow robots with exceptional cognition capabilities and reasoning skills to both understand the surrounding open world and follow natural language commands of human users~\cite{liang2023code, huang2023voxposer}. More recent works explore conversations between the human user and the robot to allow the robot to perform multi-step tasks or clarify ambiguity of the human command~\cite{wang2024mosaic, ren2023robots}. 

When the philosophy of grounding natural language commands into robot control policies is applied to human-robot collaboration (HRC), the human user may have to have a conversation with the robot at each step of the long-horizon task~\cite{wang2024mosaic}. This situation rarely happens in human-human collaboration, as a human is able to track the progress on the partner's side based on their shared knowledge over the task. For examples, a worker rarely has to have a conversation with a co-worker in a collaborative assembly task on which they have collaborated many times, and a sous chef rarely has to have a conversation with the chef when creating a regular dish together.

To address this challenge in human-robot collaboration, the robot needs to build an effective understanding of not only the environment, but also the human user. This work proposes Language-driven Intention Tracking (LIT) to model long-term behavior of the human user, and integrates LIT into an LLM-driven collaborative robot framework. LIT extends intention tracking~\cite{huang2023hierarchical} by applying an LLM to model measurement likelihood and transition probabilities in the probabilistic graphical model of human intentions, which is defined by grounding an overall task prompt (e.g., make a salad) with understanding of the scene using LLM and VLM models. Note this is the only prompt needed from the human user in LIT framework. LIT uses a VLM to generate text descriptions of the human user's behavior in the frames as measurements to track the human user's intention and filter out hallucinations. Intention prediction in the near-term allows the collaborative robot to proactively assist the human user. By harnessing the power of foundation models, we believe the LIT framework can be generalized to any collaborative tasks. We demonstrate the effectiveness of the LIT framework in a scenario where the collaborative robot acts as a sous-chef to assist a human user in cooking.

\begin{figure*}[t]
    \centering
    \includegraphics[width=\linewidth]{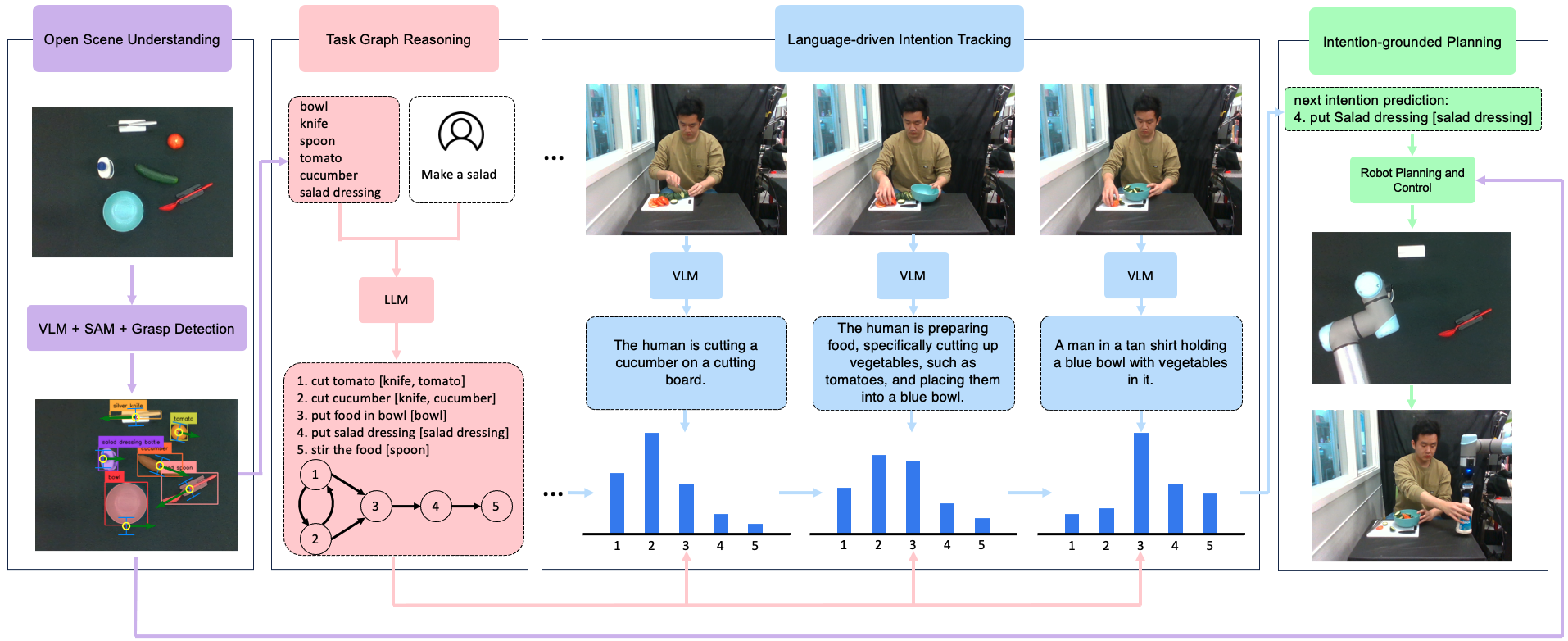}
    \caption{Language-driven Intention Tracking (LIT) based collaborative robot framework. The open scene understanding module detects objects in the scene and generate potential manipulation options, which in our case are top-down grasp poses. The task graph reasoning module takes the user's prompt on the overall task and the detected objects as input to generate a list of task steps, which we define as intention in this work. As some steps of the overall task can switch order without impact on the outcome, the LLM checks on reversibility of sequences of task steps, and builds a task graph. The Language-driven Intention Tracking module uses the task graph to build the probabilistic graphical model for intention transition. The VLM is used to generate text descriptions from frames as measurements. We compute time-varying transition probabilities and make prediction steps, and use measurements to compute measurement likelihood and make update steps to track the human intention. The intention-grounded planning module make an additional prediction step on the current intention posterior, and manipulate the objects relevant to the predicted next intention to proactively collaborate with the human.}
    \label{fig:lit-framework}
\end{figure*}
\section{Related Work}
\label{sec-related}

\subsection{Language Models in Robotics}
There has been an explosion of work in robotic planning tasks driven by LLM. 
LLMs have been shown to exhibit the ability to write multi-step control logic code based on natural language command \cite{liang2023code, pmlr-v205-huang23c}. However, such methods typically do not understand whether such a command is feasible to execute \textit{a priori}; though this can be corrected for with  e.g., feasibility planners \cite{AgiaMigimatsuEtAl2023, daf2021}, such as is done by Text2Motion \cite{Lin2023}. 
VoxPoser \cite{huang2023voxposer}, SayCan \cite{irpan2022saycan}, and LATTE \cite{latte2023} ground robotic affordances into their LLM-driven planning and modification steps following a command provided via natural language. RT-2 \cite{rt22023arxiv} extends the PALM-E VLM \cite{driess2023palme} with action representations. 
Recent works \cite{wang2023voyager, yang2023octopus} explore LLM agent embodiment within video games for accomplishing goals using visual input and/or high-level textual task description. Specifically Voyager \cite{wang2023voyager} leverages Minecraft to build an LLM-driven agent intended to continually explore, and learn new skills and store them for future use. Instead of caching skills, we propose caching intentions for more effective collaborative autonomy. 

\subsection{Intention Modeling in HRC and NLP} Understanding human intention is essential for safe and seamless human-robot interaction (HRI) and human-robot collaboration (HRC).
Human intentions have been extensively studied in various contexts, including a pedestrian's desired destination for social navigation~\cite{katyal2020intent, huang2020long, tran2021goal}, a driver's lane-changing intention for autonomous driving~\cite{kumar2013learning, xing2019driver}, and a worker's desired tool/part for collaborative manufacturing~\cite{perez2015fast, nicolis2018human}.
However, such intention estimation methods that match the observations to these well-defined intentions are usually carefully crafted and are hard to generalize to novel scenarios. On the other hand, intent classification is a crucial task in NLP and has been comprehensively studied over decades~\cite{jansen2008determining, casanueva2020efficient, minaee2021deep, firdaus2023multitask}.
In particular, identification of unseen open intents has become an emerging area in the field of intent classification~\cite{zhang2021deep, zhang2023learning}.
Nevertheless, intent classification in NLP is typically formulated as a static recognition task, whereas HRC requires using measurement sequences to perform online tracking of human intentions which can vary across time~\cite{huang2023hierarchical}.
We apply the concept of open intent in NLP to the intention tracking method for HRC, so our language-driven intention tracking method can easily generalize to novel scenarios and tasks. 

\subsection{Robotic Cooking}
There has long been interest in developing robots that can cook for and/or with people. Many works \cite{bollini2011, bollini2013, sochacki2023, shi2023a, fu2024mobile, wang2024mosaic} propose robotic systems which learn to autonomously perform cooking and baking tasks from various modalities. BakeBot \cite{bollini2011, bollini2013} and RoboCook \cite{shi2023a} take as input the plain text recipe and target shapes for deformable manipulation respectively before determining its own step-by-step instructions for the long horizon planning task of baking cookies and making dumplings from scratch. However, while the latter is able to recover from human meddling of its planned tasks, it is not able to understand why the human has interrupted the task, and will instead simply resume from a prior step to reach the original goal. Other works propose teaching a robot to cook from human demonstration~\cite{sochacki2023, fu2024mobile}. However, these works seek only to imitate the human motion of a certain cooking action, with the latter's Hidden Markov Model limited to inferring the corresponding recipe based on observations of the human skeletal motion, before subsequently continuing the recipe itself. More similar to our work, Wang \textit{et al.} propose a collaborative framework for a robotic cooking assistant in MOSAIC \cite{wang2024mosaic}, using an LLM as a task-planner and large vision models for determining locations of ingredients. However, this system requires the chef to give explicit natural language commands to incite action, and only forecasts human motion primarily as a means to provide safety for the human, such as by preventing collisions with the robot. Finally, it is specific to the cooking application.
\section{Language-driven Intention Tracking}
\label{sec:LIT}

\subsection{Problem Formulation}
A human user collaborates with a robot to perform a long-horizon multi-step task. The robot can understand the scene and reason about the task to formulate a directed task graph, where vertices are defined as steps of the task, and edges are defined as the feasible orders between the task steps. The reason we use a task graph instead of a task chain is the relationships between task steps may not always be causal and the order may be reversible, such as ``cut tomatoes'' and ``cut cucumbers'' for a salad making task. We define human intention $G_t$ as the task step the human intends to work on at time $t$. The robot uses the measurement history of the human behavior $X_{1:t}$ to track the human intention $G_t$.

\begin{figure}[hbt!]
    \centering
    \includegraphics[width=0.8\linewidth]{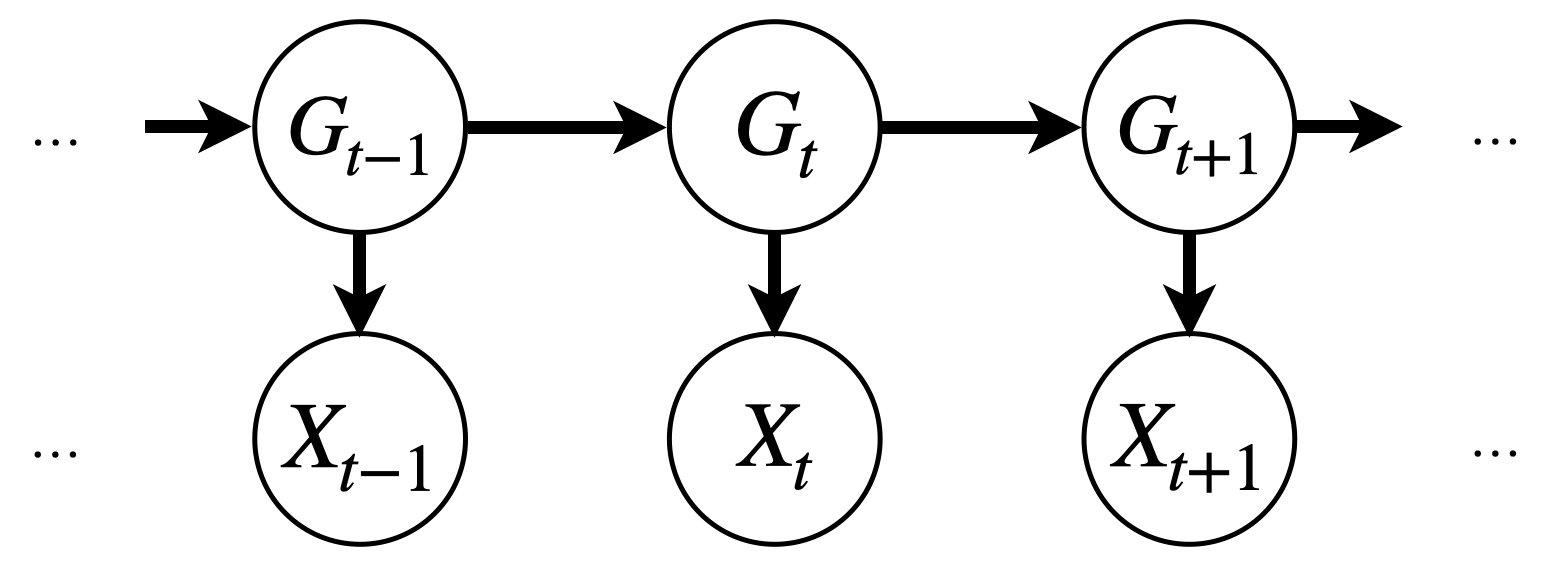}
    \caption{The graphical model for intention tracking. We denote the measurement of human behavior as $X_t$, and the human intention as $G_t$.}
    \label{fig:it-pgm}
\end{figure}

\subsection{Language Probabilistic Graphical Model}

We propose Language Probabilistic Graphical Model (LPGM) to describe the dynamics of human behavior in Fig.~\ref{fig:it-pgm}, where the value of each node is a natural language sentence. To calculate a conditional probability for example $P(A=a|B=b,C=c)$ in an LPGM, we use an LLM, where the prompt has two parts: a conditional part and a query part. We formulate the conditional part of the prompt as ``We observe \{$B$\} is \{$b$\}, and \{$C$\} is \{$c$\}, ''. We propose three different methods to compose the query part of the prompt and calculate the conditional probability accordingly.

The first method is to directly ask for $P(a|b,c)$. The query part of the prompt is formulated as ``provide the probability of \{$A$\} being \{$a$\}.''. This is similar to Ren \textit{et. al}~\cite{ren2023robots}, where the LLM is treated as an expert in modeling the mathematical relationships between $A$, $B$ and $C$. However, the outputs from the LLM may not be trustworthy if corresponding materials are not covered much by the corpus used to train the LLM. We use this as a baseline in our work.

The second method is to ask the LLM to generate a value of $A$, and compare the similarity score such as BERTScore~\cite{zhang2019bertscore} of the generated text with respect to $a$ to quantify $P(a|b,c)$. The query part of the prompt is formulated as ``what do you think \{$A$\} would be?''. This method essentially uses the LLM to provide a maximum likelihood estimate $\arg \max_{a} P(A=a|B=b,C=c)$, and uses the distance between this point estimate and the value to compute the conditional probability.

The third method addresses the case where $A$ is a discrete variable. We ask LLM to generate a list of values of $A$ with a large length $N$, compare the values in the list to all possible values of $A$, aggregate the number of most similar generated values to each possible value, and form a statistical estimate of the $P(a|b,c)$. The query part of the prompt is formulated as ``what do you think \{$A$\} would be? Provide $N$ different examples.''. This is similar to a Monte Carlo method which approximates the distribution of $P(A|B=b,C=c)$ by sampling from the LLM.

\subsection{Application in Intention Tracking}
To perform language-driven intention tracking, we iterate prediction and update steps of Bayesian filtering as presented in Eq.~\ref{eq:intention-tracking}.

\begin{equation}\label{eq:intention-tracking}
    \begin{aligned}
    & P(g_{t+1} | x_{1:t}) = \sum_{g_{t}} P(g_{t+1} | g_{t}, x_{1:t}) P(g_{t} | x_{1:t}) \\
    & P(g_{t+1} | x_{1:t+1})\! \propto\!P(x_{t+1} | x_{1:t}, g_{t+1}) P(g_{t+1} | x_{1:t}) \\
    \end{aligned}
\end{equation}

We apply LPGM methods to compute the fixed time-window approximations of intention transition $P(g_{t+1} | g_{t}, x_{t-T_w:t})$ and measurement likelihood $P(x_{t+1} | x_{t-T_w:t}, g_{t+1})$. We apply the third method to compute $P(g_{t+1} | g_{t}, x_{t-T_w:t})$, since the human intention $G_t$ is a discrete variable. We apply the second method to compute $P(x_{t+1} | x_{t-T_w:t}, g_{t+1})$.
\section{Robot Sous-Chef Application}
\label{sec:rsca}

We introduce Language-driven Intention Tracking based collaborative robot sous-chef framework as presented in Fig. \ref{fig:lit-framework}. 

\subsection{Collaborative Cooking Setup}
The human user wants to make a dish, but all the required materials and tools are not reachable by the human -- but are by the robot. The robot needs to act as a sous-chef to smoothly coordinate with the human by passing essential materials and tools at appropriate times while not making the human's cooking table overly occupied with unnecessary items at the moment. The robot is assumed to only receive the prompt at the beginning on what dish is going to be made, and will not receive prompts during collaboration.

We choose LLaVA ~\cite{liu2024llavanext, liu2023llava, liu2023improvedllava} with a 13-billion parameter Vicuna backbone \cite{vicuna2023} (derived from Llama 2 \cite{touvron2023llama}) as the VLM in the system, due to both its open source nature and competitiveness with commercial-grade models such as Gemini \cite{geminiteam2024gemini}. Note that we use the same model as the LLM for consistent performance by inputting the text prompt with a full-black image. The collaborative robot is a UR5e arm equipped with a Robotiq Hand-E Gripper. We use Intel RealSense RGBD Cameras to provide a top-down view of the robot table with objects on it, and to provide a front view of the human user's behavior. We use Robot Operating System (ROS) to build the LIT framework.

\subsection{Open Scene Understanding Module}
The overhead camera provide a top-down view of the robot workspace with all object reachable by the robot. We prompt the VLM to name and describe the objects in the image frame from the overhead camera, and take these names as input to Grounding DINO \cite{liu2023grounding} coupled with Segment Anything \cite{kirillov2023} to locate and segment all objects on the frame. Object names listed as present in the scene by the VLM that are detected with low confidence are thrown out. We perform Principal Component Analysis on object segmentations and compute object orientation and corresponding grasp poses. The detected object names and grasp poses are fed into the downstream modules.

\subsection{Task Graph Reasoning Module}
Given the available objects produced by the open scene understanding module and the general task prompt from the human user, we query the LLM to output a sequence of task steps in order to achieve the overall task.
The LLM is also asked to provide the corresponding objects needed in each task step, which will be used to inform the downstream planning module. We initialize the task graph with the sequence of task steps, and query the LLM whether adjacent steps can be reversible to add new edges to the task graph.

\subsection{Language-driven Intention Tracking Module}
We use the task steps from the reasoned task graph as possible values of intention $G_t$. The task graph is used to initialize the uniform prior among the first steps the human can start on, and to inform which pairs of $(g_t, g_{t+1})$ are required to compute the intention transition $P(g_{t+1} | g_{t}, x_{t-T_w:t})$. During collaboration, we use a front-view camera to collect frames of the human user, and feed the frames to VLM to generate text descriptions of human behavior as measurements $x_t$'s. We follow Eq.~\ref{eq:intention-tracking} to perform language-driven intention tracking.  

\subsection{Intention-grounded Planning Module}
To collaborate with the human proactively, the robot predicts the human intention at the next time step by running one prediction step in LIT on the current posterior of the intentions, and outputting the maximum probability intention as the prediction. The intention-grounded planning module performs planning and control to manipulate the objects relevant to the predicted next intention. In our collaborative cooking scenario, the robot sous-chef would pass the objects needed for the next cooking step to the human chef in advance.
\section{Preliminary Study}
\label{sec:experiments}

\begin{figure}[hbt!]
    \centering
    \includegraphics[width=\linewidth]{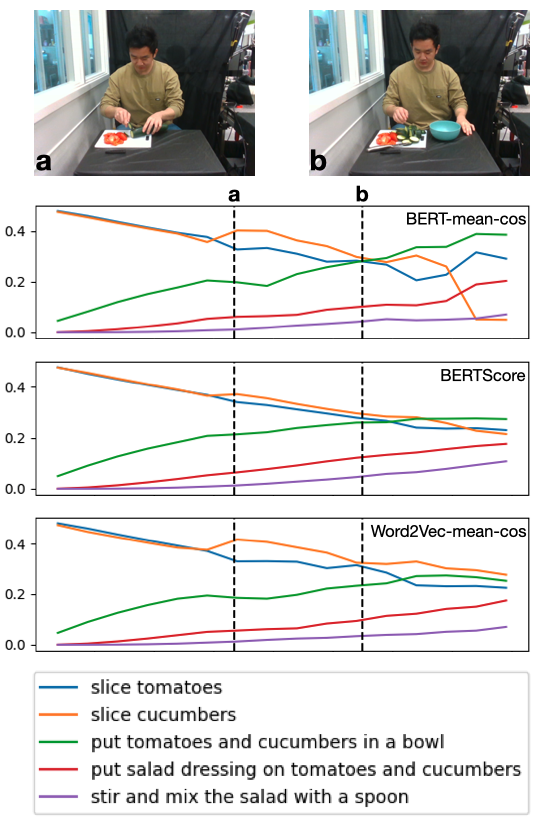}
    \caption{Language-driven Intention Tracking with different similarity metrics. The ground truth order of the human intentions: slice tomatoes; slice cucumbers; put tomatoes and cucumbers in a bowl; put salad dressing on tomatoes and cucumbers; stir and mix the salad with a spoon. Snapshots show the moment when intention transition happens. (a) The human starts cutting a cucumber after finishing cutting a tomato. (b) The human starts putting vegetables into a bowl after cutting the cucumber.}
    \label{fig:lit-result}
\end{figure}

We collect a salad cooking demonstration and run language-driven intention tracking to compare how similarity metrics affect tracking performance as presented in Fig.~\ref{fig:lit-result}. In addition to BERTScore~\cite{zhang2019bertscore}, we introduce BERT-mean-cos and Word2Vec-mean-cos~\cite{mikolov2013distributed}, which take the mean of word embeddings from the corresponding pre-trained model to generate candidate and reference sentence embeddings, and apply cosine similarity to generate similarity score. During evaluation, we isolate the effect of similarity metrics by computing measurement likelihood with the similarity metrics while using a fixed intention transition matrix based on the task graph~\cite{huang2023hierarchical}. Fig.~\ref{fig:lit-result} shows that BERT-mean-cos empirically outperforms BERTScore and Word2Vec-mean-cos for tracking human intentions.

\section{Conclusions and Future Work}
\label{sec:conclusions}
We propose Language-driven Intention Tracking (LIT) to model long-term behavior of the human user in an open scenario for proactive human-robot collaboration without repetitive prompting. We develop a LIT-based collaborative robot framework powered by LLMs and VLMs to understand the open scene, construct a task graph, track varying human intentions, and ground intention prediction to planning. We demonstrate the framework in a robot sous-chef application, where the robot seamlessly assist the human user in cooking.

In future, we will conduct human subject experiments with more comprehensive metrics to evaluate performance efficiency and user satisfaction compared to existing works. We will study the tradeoff between expressiveness and speed of the foundation models on performance of the LIT framework. We will show versatility of this framework by testing in different daily tasks such as collaborative furniture assembly. We will also work on generalization of the framework to multiple users. 

\section*{Acknowledgements}
This work was supported by the National Science Foundation under Grant No. 2143435.
  
{
    \small
    \bibliographystyle{ieeenat_fullname}
    \bibliography{main}
}


\end{document}